# A review over the applicability of image entropy in analyses of remote sensing datasets

**S.K. Katiyar** **P.V. Arun**

Department of Civil Engineering
MA National Institute of Technology, India

## Abstract

Entropy is the measure of uncertainty in any data and is adopted for maximisation of mutual information in many remote sensing operations. The availability of wide entropy variations motivated us for an investigation over the suitability preference of these versions to specific operations. The popular available versions like Tsalli's, Shannon's, and Renyi's entropies have been analysed in context of various remote sensing operations namely thresholding, clustering and registration. These methodologies have been evaluated with reference to the study area using different statistical parameters. Renyi's entropy has been found to be suitable for image registration purpose followed by Tsalli's and Shannon; whereas Tsalli's entropy has been found preferable for thresholding and clustering.

**Keywords**: entropy; clustering; thresholding; registration

## I. Introduction

Entropy as a measure of uncertainty associated with information was introduced by Shannon (1959) being inspired from the common entropy concept in physics (Long & Xin, 2000). The principle of entropy is to use uncertainty as a measure to describe the information contained in a source (Robert et al, 2003). In information theory, the concept of entropy is used to quantify the amount of information necessary to describe the macro state of a system (Frieden, 1972). Different versions of entropies have been applied for effective automation of various remote sensing analyses (Burch et al, 1983; Zhuang & Ostevold, 1987; Medha, 2009; Arun & Katiyar, 2012) and hence it is needed to investigate the suitability preference of specific entropy versions to various operations.

The entropy is related to the concept of Kolmogorov complexity, which reflects the information content of a sequence of symbols independent of any particular probability model (Zhuang & Ostevold,, 1987; Shen, 1999). The maximum information is achieved when no a priori knowledge is available, in which case, it results in maximum uncertainty. In Shannon information theory, entropy is a measure of the uncertainty over the true content of a message, but the task is complicated by the fact that successive bits in a string are not random, and therefore not mutually independent in a real message (Shannon et al, 1959). Shannon entropy is a measure of the average information content when one does not know the value of the random variable (Cover & Thomas, 1991).

Tsallis entropy (Tsallis, 1988) is obtained from the two dimensional histogram determined by using gray value of the pixels and is applied as a generalized entropy formalism (Mohamed et al, 2011). Previously, entropy has been a metric difficult to evaluate without imposing unrealistic assumptions about the data distributions (Renyi, 1960). Renyi’s entropy lends itself nicely to non-parametric estimation hence overcoming the difficulty in evaluating traditional entropy metrics (Sahoo et al, 1997). Different entropy variations can be applied for the enhancement of various remote sensing operations namely thresholding, registration and clustering; hence it is needed to investigate the suitability preference of various versions.

Thresholding is an important technique in image processing tasks and a number of methods in this context are found over the literature (Li & Vitanyi, 1997). Information theoretic approach based on the concept of entropy considers image histogram as a probability distribution, and then selects an optimal threshold value that yields the Maximum Entropy (ME) (Chang et al, 2006). ME based thresholding was first proposed by Pun (1980) and was later improved by Kapur et.al (1985). Methods suggested by Pun and Kapur are considered as first-order entropy thresholding method where as Abutaleb's (1989) cooccurrence based method and Pal's joint entropy - local entropy based method (Pal & Pal, 2003) are thought of as second-order methods. The crucial difference between entropy thresholding and relative entropy thresholding is that former maximises Shannon's entropy whereas latter minimises relative entropy. We have considered the technique proposed by Cuevas (2012) for comparative analysis.

Image registration is the process of calculating spatial geometric transforms that aligns a set of images to a common observational framework (Zitova, 2003). The feature matching step in the automation of image registration algorithms may be enhanced by adopting the maximization of mutual information. The concept of mutual information represents a measure of relative entropy between two sets, which can also be described as a measure of information redundancy (Antoine & Viergever, 1998). Mutual information enables to find an optimal match with a much better accuracy than cross-correlation, and hence improve the accuracy of registration (Wyawahare, 2009). A lot of mutual information based methods are available in literature (Elsen, 1998) and an optimal one suggested by Katiyar & Arun (2012) has been selected for comparative analysis with reference to entropy variations.

Clustering is one of the fundamental problems of pattern recognition, which organizes data patterns into natural groups or clusters in an unsupervised manner (Gokcay & Principe, 2002). Entropy can be used as similarity measure which enables to utilize all the information contained in the data distribution, and not only mere second order statistics as in many traditional algorithms (Robert et al, 2003). Groupings can be evaluated by quantifying the entropy as cluster evaluation function (CEF) which was first introduced by Gokcay & Principe (2002). The use of entropy for clustering enables to find the clusters of any shape, without knowing their true number in advance. We have adopted an entropy based clustering algorithm developed by Kin et al (2012) for comparative analysis.

In this paper we have analysed the comparative suitability of various available entropy versions in context of a few remote sensing analyses namely image thresholding, image registration and, image segmentation. The major entropies like Tsallis, Renyis and Shannon entropies were compared in the context of above operations to analyze the suitability of specific versions to specific operations.

## II. Experiment

### *2.1 Data set description*

Different satellite images of Bhopal, Chandrapur, and Kottayam with one year variation (i.e. 2011 & 2012,) have been taken as test images for comparing performances of various entropy enhanced implementations. The details of the satellite data adopted for these investigations are summarised in (Table 1) and details of ground truthing survey is presented in (Table 2).

Table 1. Data description

| ***S.No*** | ***Satellite*** | ***Sensor*** | ***Date of Procurement*** | ***Resolution(m)*** |
|---|---|---|---|---|
| 1 | IRS-P5 | LISS 4 | November,2012, November,2011 | 5.8 |
| 2 | IRS-P5 | LISS3 | November,2012 November,2011 | 23.5 |
| 3 | LANDSAT | TM | November,2012 November,2011 | 30 |
| 4 | CARTOSAT-II | PAN | November,2012 November,2011 | 2.5 |

Table 2. Ground truthing information

| S.No | Area | Date of procurement | No. Classes | No .of points / class |
|---|---|---|---|---|
| 1 | Kottayam | August, 2012, | 5 | 30 |
| 2 | Bhopal | November,2012, | 5 | 40 |
| 3 | Chandrapur | November,2012, | 5 | 35 |

### *2.2 Methodology: Comparative analysis of different algorithms*

Literature review has been conducted to select accurate and recent methodologies in the context of thresholding, registration and clustering. These algorithms were modified to enhance them with entropies and to facilitate the variation of entropy. The various algorithms were implemented in MATLAB and the accuracies were computed using ERDAS imagine and the results were compared. Ground truthing for various accuracy estimations were done with reference to the Google earth and Differential Global Positioning System (DGPS) survey data over the Bhopal city and surrounding areas. The methodology adopted for this research work is as given in (Figure 1).

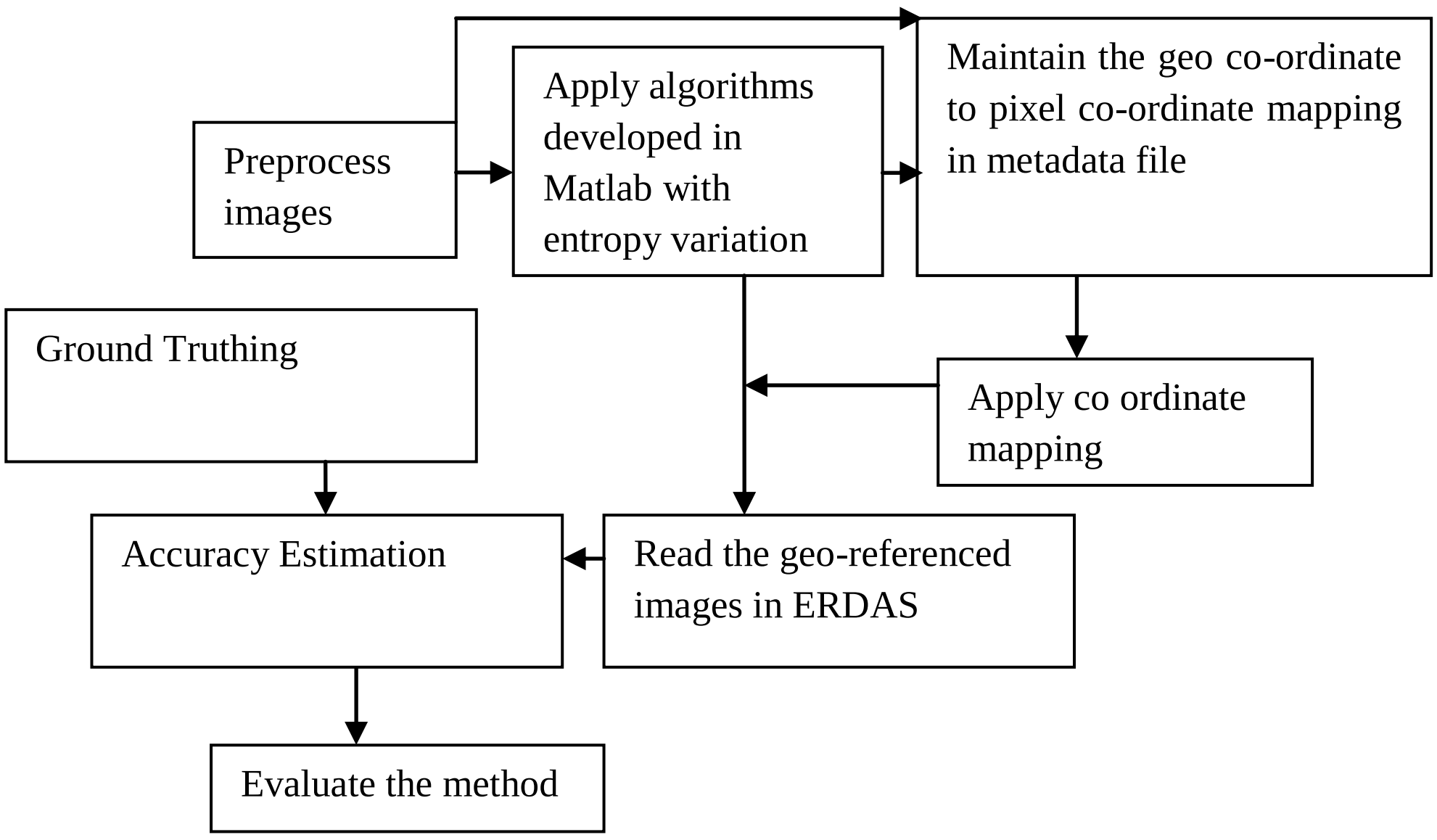

Figure 1. Methodology Adopted

### *2.2.1 Pre-processing*

Our aim has been to generalise the results and hence we have chosen different locations with varied landscape complexity. Pre-processing has been done to remove the salt and pepper noises without much altering the original entropies. Hence the minor alteration of entropies may be justified as we have been aiming over noise removal. The images have been collected over same season, ie during November for Bhopal, October for Chandrapur and during September for Kottayam to avoid much seasonal effects.

### *2.2.2 Registration*

Suitability of entropy versions for registration process have been analysed by adopting the method developed by Arun & Katiyar (2012) and accuracy is estimated using Normalized Cross Correlation coefficient (NCCC) (Wyawahare, 2009) and Root Mean Square Error (RMSE) (Antoine & Viergever, 1998).

### *2.2.3 Segmentation*

The comparative analysis of the different entropies in context of image segmentation have been analysed based on method developed by Kim et.al (2012). Different statistical parameters namely kappa statistics and over all accuracy have been used for accuracy comparison.

### *2.2.4 Thresholding*

Investigations over the suitability of entropy variations for multi level thresholding have been carried out based on the thresholding algorithm suggested by Cuevas (2012). Proposed algorithm has been modified to include cross entropy criterion for improving the results. Thresholding is adopted for segmentation of images and test is carried out at various thresholding levels namely 2- 5. The various levels of thresholding over different satellite images showed similar trends in accuracy evaluation. The accuracy of image thresholding implementations has been evaluated using kappa statistics (Elsen, 1998) and overall accuracy (Mehmet & Bu, 2004).

## III. Results and discussion

### *3.1. Registration*

The comparative analysis of various entropies over image registration techniques have been verified on various satellite images as LISS III, LISS IV and the results of observations are as summarised in the (Table 3). Normalized Cross Correlation Coefficient (NCCC), Root Mean Square Error (RMSE) and execution time were used as evaluation criteria. NCCC measures the similarity between the images with values ranging from (0-1) and an NCCC value of unity indicates perfectly registered images. RMSE value indicates error in registration and a least RMSE value is preferred for a perfect registration. The execution time was also analyzed using MATLAB counter function and generally categorized as high (>60 sec), medium (30-60 sec), low (<30 sec). RMSE value is lowest and NCCC is highest for Renyis followed by TSalli's in majority cases. These results in Table 3 confirms that the former is more preferable followed by latter when compared to the Shannon.

Table 3. Accuracy comparison of registration

| S.No | TEST DATA Master Image | TECHNIQUE | STUDY AREA | NCCC | RMSE | EXECUTION TIME |
|---|---|---|---|---|---|---|
| 1 | LISS3 | Tsallis | Bhopal | 0.65 | 5.13 | Medium |
| | | | Chandrapur | 0.68 | 4.12 | Medium |
| | | | Kottayam | 0.71 | 3.86 | Medium |
| | | Renyi's | Bhopal | 0.67 | 3.93 | Low |
| | | | Chandrapur | 0.76 | 3.81 | Low |
| | | | Kottayam | 0.81 | 3.52 | Low |
| | | Shannon | Bhopal | 0.53 | 5.85 | Higher |
| | | | Chandrapur | 0.61 | 4.82 | Higher |
| | | | Kottayam | 0.67 | 4.18 | Higher |
| 2 | Google Earth | Tsallis | Bhopal | 0.71 | 3.82 | High |
| | | | Chandrapur | 0.74 | 3.31 | High |
| | | | Kottayam | 0.76 | 3.28 | High |
| | | Renyi's | Bhopal | 0.70 | 3.12 | Low |
| | | | Chandrapur | 0.78 | 2.85 | Low |
| | | | Kottayam | 0.81 | 2.23 | Low |
| | | Shannon | Bhopal | 0.69 | 4.82 | Higher |
| | | | Chandrapur | 0.71 | 4.27 | Higher |
| | | | Kottayam | 0.72 | 4.08 | Higher |
| | | Tsallis | Bhopal | 0.65 | 5.72 | Low |
| | | | Chandrapur | 0.67 | 5.19 | Low |
| | | | Kottayam | 0.71 | 4.37 | Low |
| 3 | LANDSAT | Renyi's | Bhopal | 0.69 | 4.12 | Low |
| | | | Chandrapur | 0.71 | 4.03 | Low |
| | | | Kottayam | 0.76 | 4.15 | Low |
| | | Shannon | Bhopal | 0.53 | 6.01 | Higher |
| | | | Chandrapur | 0.59 | 5.71 | Higher |
| | | | Kottayam | 0.63 | 5.51 | Higher |
| 4 | LISS4 | Tsallis | Bhopal | 0.78 | 3.94 | High |
| | | | Chandrapur | 0.77 | 3.71 | High |
| | | | Kottayam | 0.73 | 3.62 | High |
| | | Renyi's | Bhopal | 0.81 | 3.81 | Medium |
| | | | Chandrapur | 0.84 | 3.52 | Medium |
| | | | Kottayam | 0.85 | 3.48 | Medium |
| | | Shannon | Bhopal | 0.69 | 4.32 | Higher |
| | | | Chandrapur | 0.68 | 4.18 | Higher |
| | | | Kottayam | 0.71 | 3.96 | Higher |

### *3.2. Segmentation*

Investigations on the suitability of entropy versions for image segmentation revealed that Tsallis entropy is preferable followed by Renyi's and Shannon. Efficiency of entropy variations are evaluated with reference to different satellite images of the study areas. Comparative accuracies in terms of various statistical measures are summarised in (Table 4). Increase in value of kappa statistics (maximum = 1) and overall accuracy (max=100%) indicates better methodology.

Table 4. Accuracy Comparison

| S.No | Sensor | Methodology (Entropy adopted) | Area | Kappa statistics | Overall Accuracy (%) |
|---|---|---|---|---|---|
| 1 | LISS 3 | Tsallis | Bhopal | 0.96 | 96.83 |
| | | | Chandrapur | 0.94 | 94.32 |
| | | | Kottayam | 0.91 | 91.10 |
| | | Renyi's | Bhopal | 0.93 | 94.58 |
| | | | Chandrapur | 0.89 | 93.68 |
| | | | Kottayam | 0.84 | 91.01 |
| | | Shannon | Bhopal | 0.92 | 93.13 |
| | | | Chandrapur | 0.86 | 90.03 |
| | | | Kottayam | 0.81 | 89.01 |
| 2 | LISS 4 | Tsallis | Bhopal | 0.92 | 94.80 |
| | | | Chandrapur | 0.91 | 92.10 |
| | | | Kottayam | 0.90 | 89.78 |
| | | Renyi's | Bhopal | 0.85 | 93.00 |
| | | | Chandrapur | 0.82 | 92.48 |
| | | | Kottayam | 0.79 | 91.87 |
| | | Shannon | Bhopal | 0.88 | 91.00 |
| | | | Chandrapur | 0.86 | 88.07 |
| | | | Kottayam | 0.83 | 87.91 |
| 3 | LANDSAT | Tsallis | Bhopal | 0.75 | 95.80 |
| | | | Chandrapur | 0.72 | 93.11 |
| | | | Kottayam | 0.68 | 92.68 |
| | | Renyi’s | Bhopal | 0.71 | 93.00 |
| | | | Chandrapur | 0.69 | 92.23 |
| | | | Kottayam | 0.63 | 91.21 |
| | | Shannon | Bhopal | 0.65 | 91.08 |
| | | | Chandrapur | 0.61 | 88.23 |
| | | | Kottayam | 0.56 | 87.23 |
| 4 | Google Earth | Tsallis | Bhopal | 0.90 | 94.80 |
| | | | Chandrapur | 0.88 | 89.12 |
| | | | Kottayam | 0.81 | 88.34 |
| | | Renyi’s | Bhopal | 0.83 | 87.65 |
| | | | Chandrapur | 0.80 | 83.56 |
| | | | Kottayam | 0.76 | 83.90 |
| | | Shannon | Bhopal | 0.81 | 81.34 |
| | | | Chandrapur | 0.73 | 82.56 |
| | | | Kottayam | 0.72 | 82.34 |

The visul results of segmentation using different entropy techniques is as presented in (Figure 2) which shows that Tsalli’s should be preferred followed by Renyi’s and Shannon.

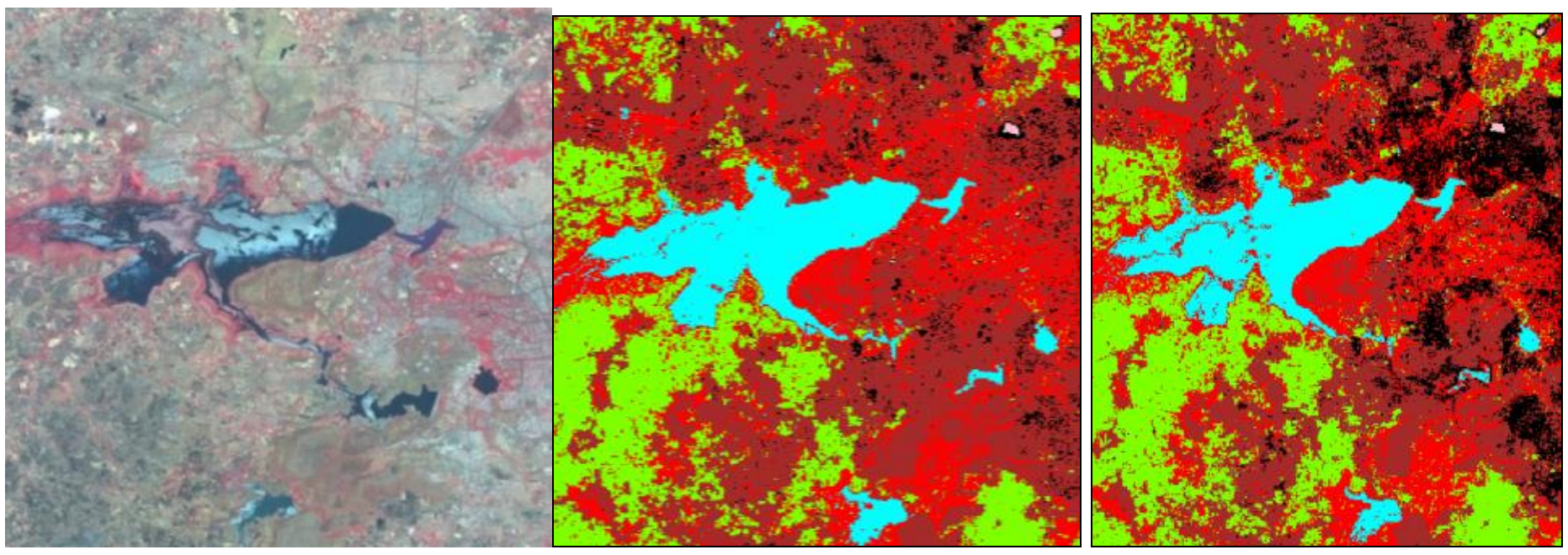

(a) LISS 3 (b) Shannon (c) Renyis

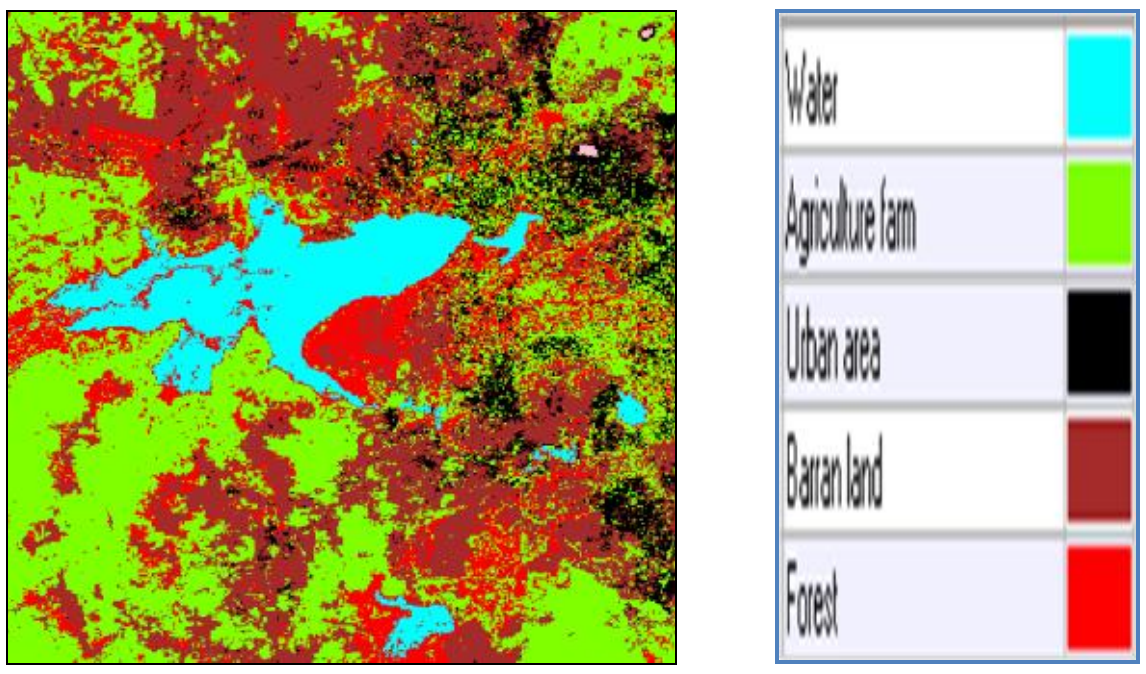

(d) Tsallis (e) Index

Figure 2. Visual comparison of different segmentation on LISS3 sensor imagery

### *3.3. Thresholding*

Effect of variable levels of multi thresholding with reference to different entropy variations are presented in (Table 3). Comparative analysis of entropy techniques in the context of thresholding revealed that Tsallis entropy is comparatively more suitable when compared to Renyi's and Shannon. The results of multi level thresholding (level-5) are summarised in (Table 5). Kappa value and overall efficiency have been used for accuracy estimation where a higher value of both corresponds to better technique.

Table 5. Accuracy Comparison

| S.No | Sensor | Methodology (Entropy adopted) | Area | Kappa statistics | Over all Accuracy |
|---|---|---|---|---|---|
| 1 | LISS 3 | Tsallis | Bhopal | 0.85 | 89.68 |
| | | | Chandrapur | 0.82 | 85.28 |
| | | | Kottayam | 0.80 | 82.12 |
| | | Renyi's | Bhopal | 0.82 | 87.98 |
| | | | Chandrapur | 0.77 | 81.23 |
| | | | Kottayam | 0.73 | 73.57 |
| | | Shannon | Bhopal | 0.81 | 73.34 |
| | | | Chandrapur | 0.83 | 68.46 |
| | | | Kottayam | 0.68 | 64.09 |
| 2 | LISS 4 | Tsallis | Bhopal | 0.81 | 78.13 |
| | | | Chandrapur | 0.78 | 71.66 |
| | | | Kottayam | 0.79 | 75.12 |
| | | Renyi's | Bhopal | 0.74 | 74.59 |
| | | | Chandrapur | 0.70 | 75.48 |
| | | | Kottayam | 0.68 | 72.23 |
| | | Shannon | Bhopal | 0.77 | 68.56 |
| | | | Chandrapur | 0.72 | 64.24 |
| | | | Kottayam | 0.66 | 63.20 |
| 3 | LANDSAT | Tsallis | Bhopal | 0.63 | 81.09 |
| | | | Chandrapur | 0.59 | 78.23 |
| | | | Kottayam | 0.56 | 76.12 |
| | | Renyi's | Bhopal | 0.58 | 78.25 |
| | | | Chandrapur | 0.54 | 75.89 |
| | | | Kottayam | 0.49 | 73.09 |
| | | Shannon | Bhopal | 0.51 | 74.56 |
| | | | Chandrapur | 0.48 | 71.89 |
| | | | Kottayam | 0.43 | 69.34 |
| | Google | Tsallis | Bhopal | 0.88 | 68.80 |
| | | | Chandrapur | 0.76 | 65.80 |
| 4 | Earth | | Kottayam | 0.72 | 63.45 |
| | | Renyi's | Bhopal | 0.70 | 64.88 |
| | | | Chandrapur | 0.68 | 61.98 |
| | | | Kottayam | 0.65 | 59.41 |
| | | Shannon | Bhopal | 0.69 | 62.05 |
| | | | Chandrapur | 0.61 | 56.53 |
| | | | Kottayam | 0.60 | 53.12 |

The performance of different entropies with reference to various levels (1-5) of thresholding for LISS 4 imagery and LANDSAT imageries are presented in Figure 3 & 4 respectively. Investigation over all sample imageries have shown similar trend and is evident form graphical representation of analysis over finest and coarsest imageries of our data sets.

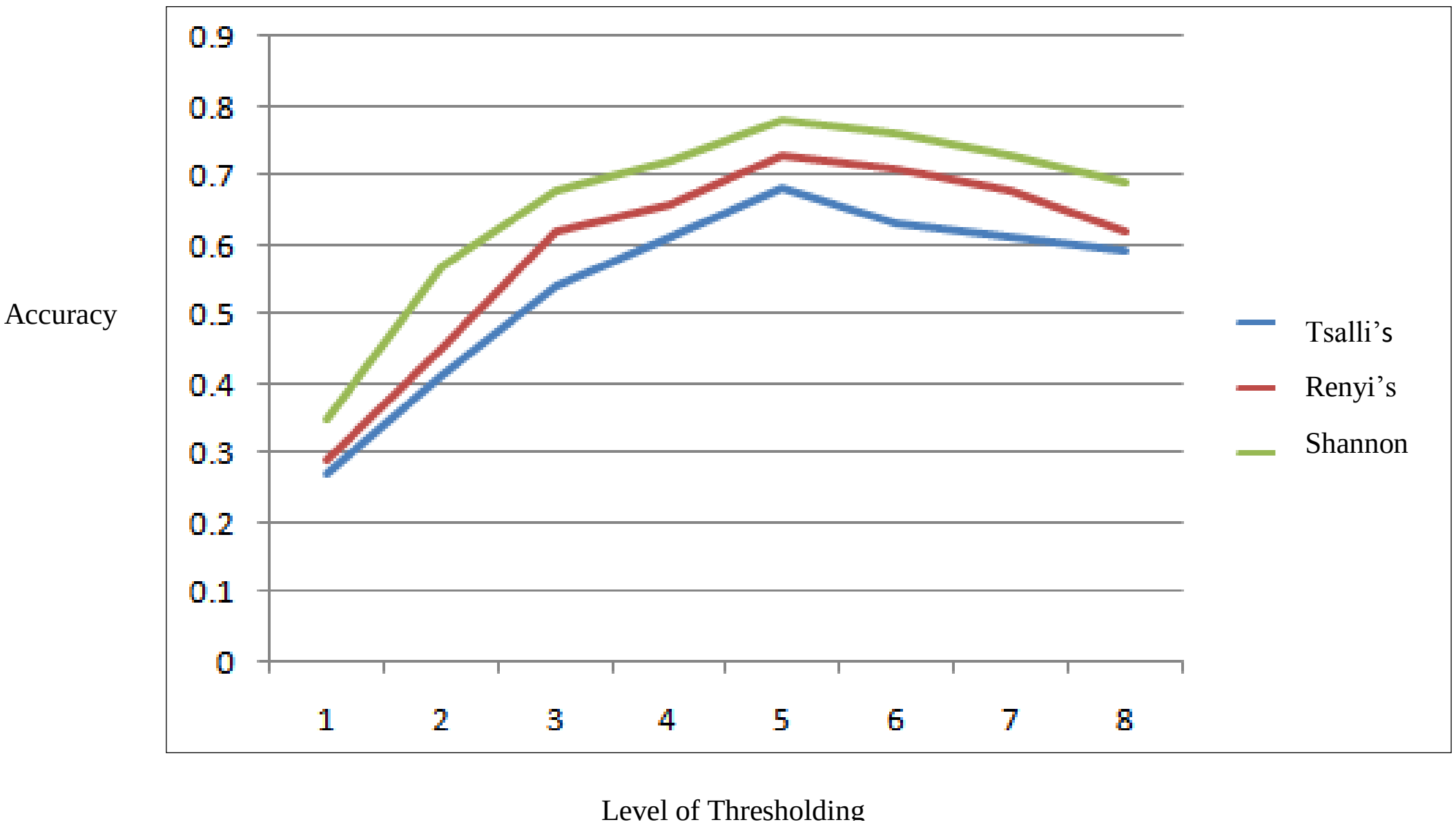

Figure 3. Accuracy variation of different entropies for different levels of thresholding

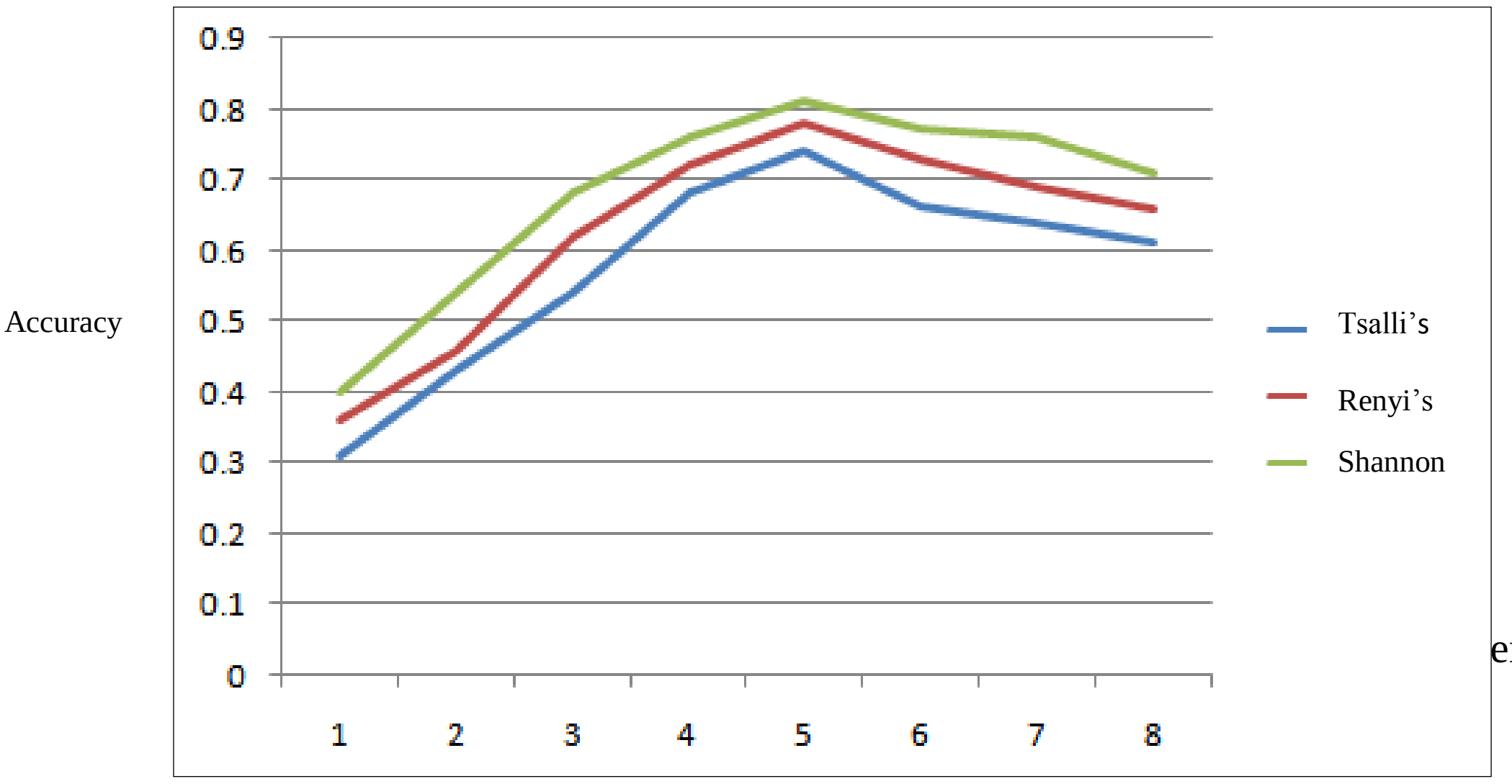

ensing

Level of Thresholding

Figure 4. Accuracy variation of different entropies for different levels of thresholding

Figure 4 also reveals that the accuracy increases with increase in threshold level however drops back after an optimal threshold (in this case 5) which is the characteristic of the scene. Visual results of thresholding over LISS 4 imagery is presented in (Figure 5) which shows a considerable improvement using Tsalli's when compared to Renyi's and Shannon.

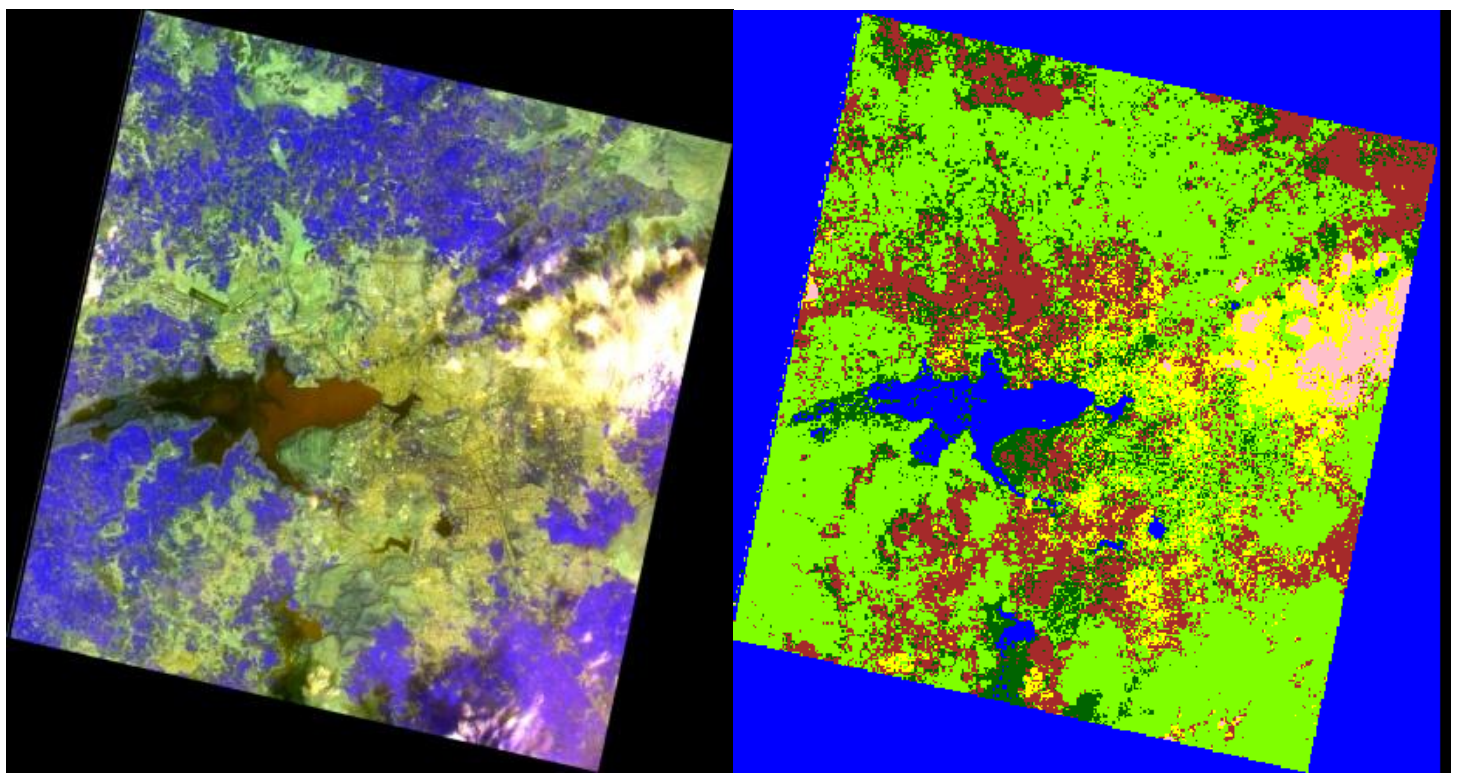

a)LISS 4 b)Shannon

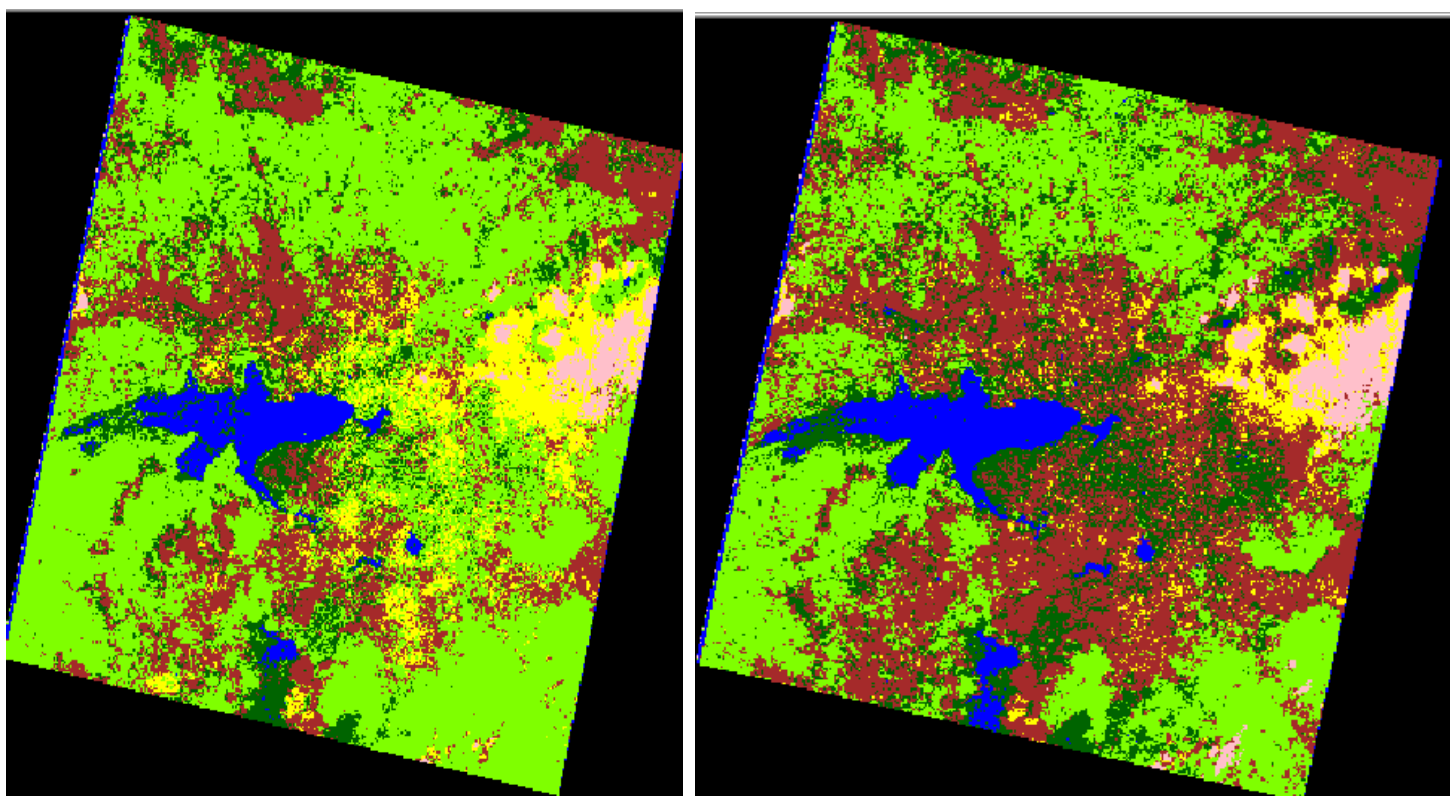

c)Renyi's d)Tsalli's

Figure 5. Visual comparison of thresholding methods on LISS4 sensor imagery

## IV. Conclusion

The investigations over the suitability of different entropy techniques in the context of various remote sensing operations revealed that Renyi's entropy is suitable for image registration purpose followed by Tsalli's and Shannon where as Tsalli's entropy is found preferable for thresholding and clustering. Renyi's is the simplest entropy method and is computationally simple as it avoids parametric estimation. The various experiments revealed that Shannon is the computationally complex entropy variation and hence is less preferable in different operations.